\documentclass[lettersize,journal]{IEEEtran}
\usepackage{amsmath,amsfonts}
\usepackage{algorithmic}
\usepackage{algorithm}
\usepackage{array}
\usepackage[caption=false,font=normalsize,labelfont=sf,textfont=sf]{subfig}
\usepackage{textcomp}
\usepackage{stfloats}
\usepackage{url}
\usepackage{verbatim}
\usepackage{graphicx}
\usepackage{cite}
\usepackage{booktabs}
\usepackage{multirow}
\usepackage{color}

\begin{document}

\title{GroupTransNet: Group Transformer Network for RGB-D Salient Object Detection}

\author{Xian Fang, Jinshao Zhu, Xiuli Shao, and Hongpeng Wang$^*$,~\IEEEmembership{Member,~IEEE}

\thanks{Xian Fang and Xiuli Shao are with the College of Computer Science, Nankai university, Tianjin 300350, China. Jinchao Zhu and Hongpeng Wang are with the College of Artificial Intelligence, Nankai university, Tianjin 300350, China.}
\thanks{$^*$ Corresponding author: Hongpeng Wang (e-mail: hpwang@nankai.edu.cn)}}

\markboth{Journal of \LaTeX\ Class Files,~Vol.~14, No.~8, August~2021}%
{Shell \MakeLowercase{\textit{et al.}}: A Sample Article Using IEEEtran.cls for IEEE Journals}


\maketitle

\begin{abstract}
Salient object detection on RGB-D images is an active topic in computer vision, and has witnessed substantial progress. Although the existing methods have achieved appreciable performance, there are still some challenges. The locality of convolutional neural network requires that the model has a sufficiently deep global receptive field, which always leads to the loss of local details. To address the challenge, we propose a novel Group Transformer Network (GroupTransNet) for RGB-D salient object detection. This method is good at learning the long-range dependencies of cross layer features to promote more perfect feature expression. At the beginning, the features of the slightly higher classes of the middle three levels and the latter three levels are soft grouped to absorb the advantages of the high-level features. The input features are repeatedly purified and enhanced by the attention mechanism to purify the cross modal features of color modal and depth modal. The features of the intermediate process are first fused by the features of different layers, and then processed by several transformers in multiple groups, which not only makes the size of the features of each scale unified and interrelated, but also achieves the effect of sharing the weight of the features within the group. The output features in different groups complete the clustering staggered by two owing to the level difference, and combine with the low-level features.
Extensive experiments demonstrate that GroupTransNet outperforms the comparison models and achieves the new state-of-the-art performance.
\end{abstract}


\section{Introduction}

\IEEEPARstart{T}{he} goal of RGB-D salient object detection is to identify and segment the most eye-catching region or target in a pair of cross-modal RGB image and depth image. It has been widely used in many computer vision tasks, such as object detection \cite{Felzenszwalb2010Object, Wang2014Moving}, visual tracking \cite{Hong2015Online, Jia2016Visual} and image retrieval \cite{Shao2006Specific, Radenovic2019Fine-tuning}, and serves them as the downstream subtasks. At present, with the popularity of digital cameras, smartphones and other imaging devices, the acquisition of depth image becomes growing convenient. In this situation, only using a single color image is gradually difficult or unable to meet the needs of advanced detection. In fact, color images can give color clues to obtain the texture information of different objects. At the same time, depth images can record the distance information of different objects and provide additional clues to capture the spatial structure and three-dimensional layout. Introducing depth information together with color information into salient object detection is helpful to solve some thorny dilemmas.

Approaches such as early fusion, late fusion and cross level fusion have been proposed to focus on the complementarity of color modal and depth modal. They first separately extract the RGB and depth features, and then fuse the shallow, deep and cross layer features of the constructed network, respectively. Most of the existing methods use Convolutional Neural Networks (CNNs) based on U-Net \cite{Ronneberger2015U-Net} to encode and decode the features of different layers in deep and shallow layers. Among them, the high-level features from the deep layer are rich in rough semantic information, while the low-level features from the shallow layer are rich in precise detail information. Such a mode is consistent with the basic mode of encoder-decoder, that is, the encoder first extracts features from the input image, and then the decoder aggregates features to predict the saliency map. Most recently, the transformer has caused an uproar in computer vision, because it can effectively model the global long-range dependencies of patches in the data expression of images. A natural legal assumption is to leverage a transformer to replace or combine CNNs for salient object detection. Although it is very effective to simply use the former to replace the latter, the integration of the two will be a better option.

Inspired by TriTransNet \cite{Liu2021TriTransNet}, it is observed that transformer encoder with shared weight can better obtain the common information of multiple features. Different from the previous methods, we propose a Group Transformer Network (GroupTransNet). To the best of our knowledge, GroupTransNet is the first attempt to express the futures of transformer encoder in the data structure of the group. The energy weights in the same group pursue the consistency of features, while the energy weights outside different groups pursue the difference of features, which makes the mixed features more recognizable. GroupTransNet consists of four components, i.e., modal purification module, scale unification module, multiple transformer encoder and cluster integration unit. Firstly, the cross modal features of color mode and depth mode are purified and enhanced by the modal purification module. Then, the scale unification module coordinates the size and interaction of each scale feature. After that, the multiple transformer encoder learns the long-range dependencies of features and ensures high cohesion and low coupling in different groups. Finally, the ideal features are output in cascade by the cluster integration unit.

In summary, our major contributions are three folds:
\begin{itemize}
  \item A modal purification module and a scale unification module are proposed, which purify cross modal features through repeated purification and attention enhancement, and ensure that the size of each scale feature is unified and interrelated through the superposition and fusion of upsampling and downsampling, respectively.
  \item A multiple transformer encoder is proposed, in which the grouped transformer bodies with multiple shared weights can better and more specifically express the features, and learn the common information of cross modal and scale features within different groups.
  \item A cluster integration unit is proposed, which uses the dislocation deviation of the level of features in the group to realize the robust combination of different types of high-level features to low-level features.
  \item Comprehensive experiments on six benchmark datasets under four typical evaluation metrics demonstrate that the proposed GroupTransNet is very competitive to 18 state-of-the-art methods.
\end{itemize}

The rest of this paper is organized as follows. Section \uppercase\expandafter{\romannumeral2} briefly surveys some related works. Section \uppercase\expandafter{\romannumeral3} describe our proposed method in detail. Extensive experiments are conducted in Section \uppercase\expandafter{\romannumeral4}. The conclusion is given in Section \uppercase\expandafter{\romannumeral5}.

\section{Related Works}
In this section, we present a brief review of the works related to this paper from two aspects, including saliency detection and transformer.

\subsection{Saliency detection}

Saliency object detection is called saliency detection for short. So far, researchers have proposed a large number of saliency detection methods. These methods mainly design models to aggregate high-level and low-level features.
For instance,
Wu et al. \cite{Wu2019Stacked} utilizes the interaction between the edge module and the detection module to optimize these two tasks at the same time.
Su et al. \cite{Su2019Selectivity} uses boundary to guide encoder and decoder to gradually optimize saliency prediction.
Chen et al. \cite{Chen2020Global} integrates low-level appearance features, high-level semantic features and global context features, and generate saliency map in a supervised manner.

In recent years, it has been found that the depth information with three-dimensional layout and spatial structure can assist the color information, and the use of these two kinds of information can improve the detection performance. Therefore, researchers have proposed plenty of RGB-D saliency detection methods. According to the fusion form of the two modals, these methods can be divided into early fusion, late fusion and cross level fusion.
For instance,
Qu et al. \cite{Qu2017RGBD} proposes an interactive mechanism for automatic learning RGB-D saliency object detection.
Zhang et al. \cite{Zhang2020UC-Net} proposes to simulate the uncertainty of human annotation through the network of conditional variational automatic encoder, and generate multiple saliency maps for each input image by sampling in the potential space.
Wang et al. \cite{Wang2019Adaptive} designs two stream convolution neural networks, in which each network extracted features from color or depth patterns and predicted the significance map, and learned the switch map to adaptively fuse the predicted saliency map.
Piao et al. \cite{Piao2020A2dele} realizes the expected control of pixel level depth knowledge transmitted to the color stream by adaptively minimizing the difference between the prediction generated by the depth stream and the color stream, and transformed the positioning knowledge into color features.
Pang et al. \cite{Pang2020Hierarchical} integrates the features of different modals through densely connected structures, and use their mixed features to generate dynamic filters with different sizes of receptive fields.
Ji et al. \cite{Ji2020Accurate} proposes a new collaborative learning framework to make use of edge, depth and saliency in a more effective way.
Fan et al. \cite{Fan2021Rethinking} proposes to realize the filtering of low-quality depth map and cross modal feature learning.

\subsection{Transformer}

Transformer was first proposed by \cite{Vaswani2017Attention}. Once proposed, it quickly occupies a dominant position in Natural Language Processing (NLP), which is used to model global long-range dependencies, constantly refreshing records one after another. Recently, thanks to its success, researchers have extended it into computer vision and achieved impressive results, thus steadily winning a place.

For instance,
Liu et al. \cite{Liu2021Swin1} proposes a hierarchical transformer with shift window scheme.
Liu et al. \cite{Liu2021Swin2} proposes to make the model transmit more effectively across window resolution.
Chen et al \cite{Chen2021TransUNet} proposes using transformer to encode the labeled image block in CNNs feature map into an input sequence for extracting global context. At the same time, the decoder is used to upsample the encoded features and combine them with high-resolution CNNs feature map to achieve accurate positioning.
Dosovitskiy et al \cite{Dosovitskiy2021An} interprets the image as a sequence of flat two-dimensional patches, and uses transformer for image classification.
Yuan et al \cite{Yuan2021Tokens-to-Token} uses T2T module to model the local structure, so as to generate multi-scale token features.
Liu et al \cite{Liu2021Visual} takes image patches as input and used transformer to propagate global context between image patches.
Han et al \cite{Han2021Transformer} regards local patches as visual sentences and further divided them into smaller patches as visual words. The attention of each word will be calculated together with other words in a given visual sentence.
Li et al \cite{Li2021LocalViT} improves the locality mechanism of information exchange in the local area by introducing deep convolution into the feedforward network and adding locality to the visual transformer.
Liu et al \cite{Liu2021TriTransNet} proposes the triple transformer embedded module to learn cross layer long-range dependencies to enhance high-level features.
Tang et al. \cite{Tang2021CoSformer} proposes to capture significant and common visual patterns from multiple images.
Ren et al \cite{Ren2021Unifying} proposes a pure transformer based encoder and a hybrid decoder to aggregate the features generated by transformer.
Wang et al \cite{Wang2021MTFNet} respectively introduces a new pixel level focus regularization to guide CNNs feature extraction program, and deeply utilizes the multimodal interaction of color and depth images on coarse and fine scales.

\begin{figure*}[t] \small
  \centering
  \includegraphics[width=0.99\textwidth]{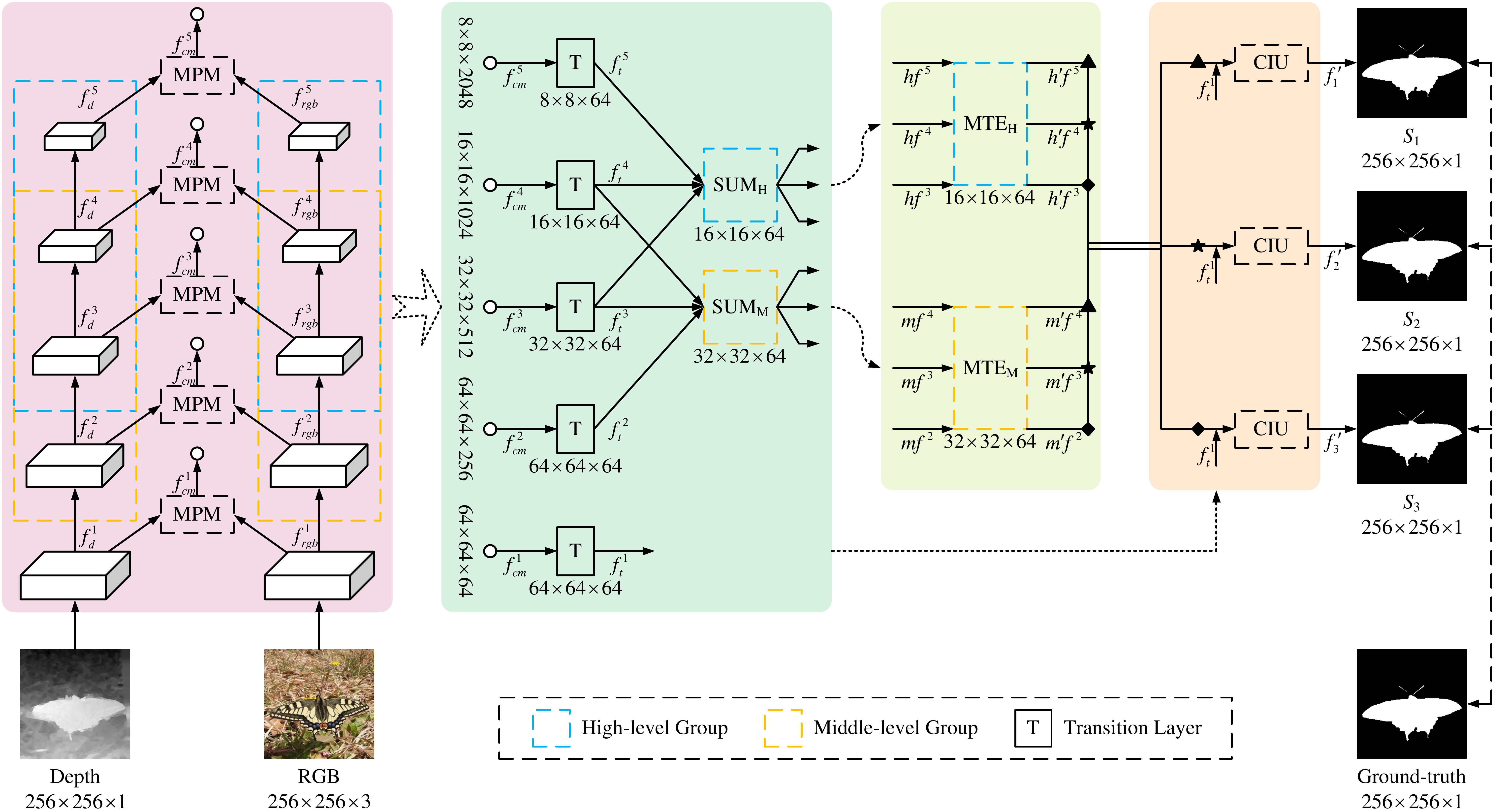} \\
  \caption{The overall architecture of our Group Transformer Network (GroupTransNet).}
  \label{Figure 1}
\end{figure*}

\section{Proposed Method}

In this section, we describe the proposed Group Transformer Network (GroupTransNet) in detail. Specifically, we firstly show the network architecture, then introduce the proposed components in turn, and finally give the loss function.

\subsection{Architecture Overview}
As shown in Figure \ref{Figure 1}, the overall architecture of the proposed GroupTransNet consists of Modal Purification Module (MPM), Scale Unification Module (SUM), Multiple Transformer Encoder (MTE) and Cluster Integration Unit (CIU).
In the whole process, for these different scale features $f^{1}$, $f^{2}$, $f^{3}$, $f^{4}$ and $f^{5}$ from color modal and depth modal, they are soft divided into two groups, $G_{1} = \{f^{2}, f^{3}, f^{4}\}$ and $G_{2} = \{f^{3}, f^{4}, f^{5}\}$. First, all input features are purified by MPM to obtain cross modal features. Second, the grouped features and are processed by SUM to make the size of each scale feature unified and interrelated. Third, the intermediate features learn the common information in these groups pass through MTE to get more discriminative feature representation. Fourth, the combination of high level features to low level features to generate output features, which is realized in CIU. The final saliency map is computed by integrating the feature map obtained.

\subsection{Modal Purification Module}
In RGB-D saliency detection, there are two kinds of materialized expression forms of image information, which are color modal and depth modal. Specifically, the color modal provides the appearance clues of the image, while the depth modal gives the distance clues of the image. They add useful information for detection in different modes. However, the features of different modals are incompatible to a certain extent, which is caused by the inherent differences of modals. If we simply utilize these two modals, such as concatenating the color modal directly with the depth modal, it will bring serious noise to the features. To address this, a Modal Purification Module (MPM) is designed.

\begin{figure}[!t] \small
  \centering
  \includegraphics[width=0.34\textwidth]{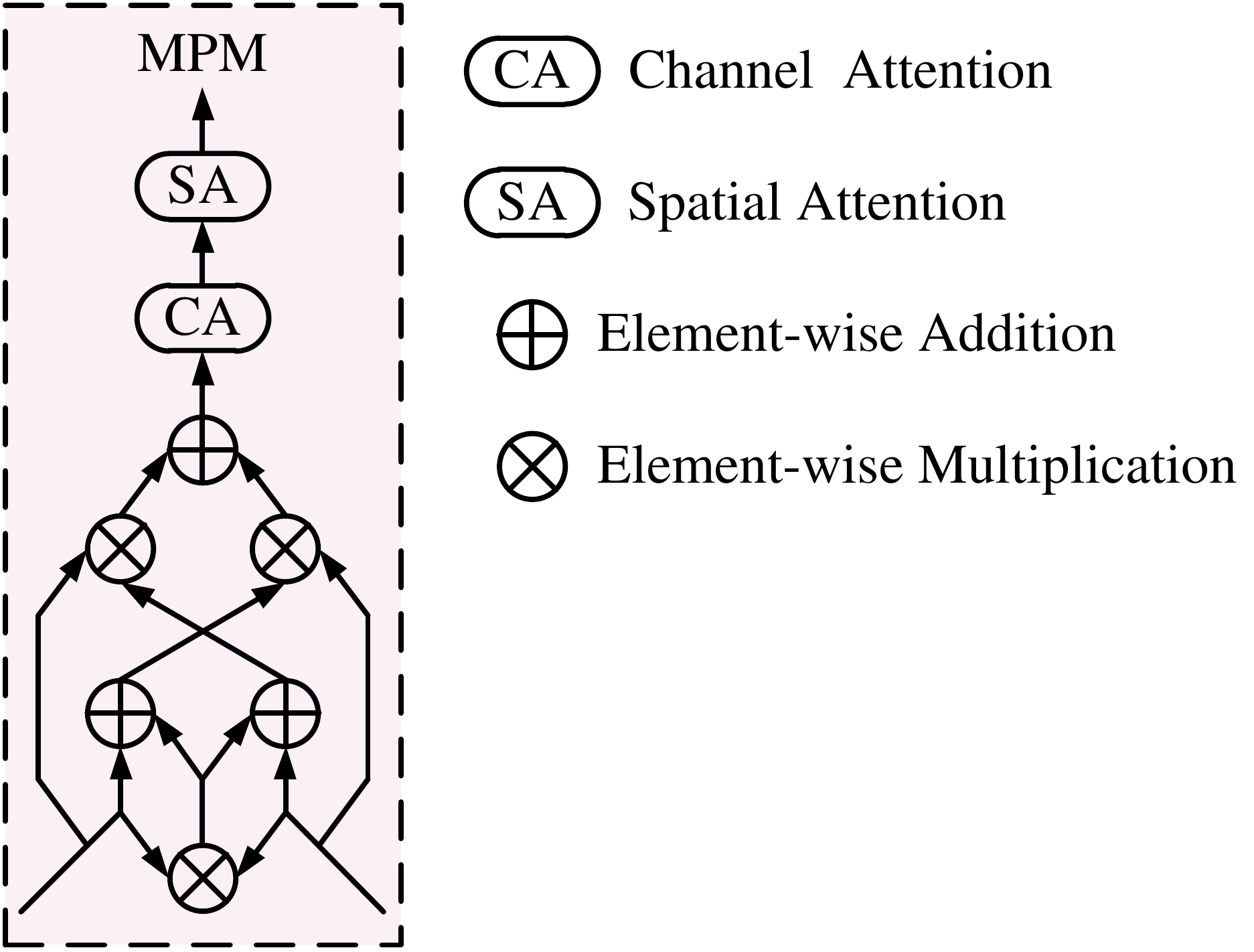} \\
  \caption{The schematic diagram of Modal Purification Module (MPM).}
  \label{Figure 2}
\end{figure}

The schematic diagram of MPM is shown in Figure \ref{Figure 2}. MPM first carries out feature purification by repeated element level linkage, and then carries out feature enhancement by attention mechanism. Among them, the attention mechanism refers to channel attention mechanism and spatial attention mechanism, which are stated in CBAM \cite{Woo2018CBAM}.

For color feature $f_{rgb}^{i}$ $(i=1,2,3,4,5)$ and depth feature $f_{d}^{i}$ $(i=1,2,3,4,5)$, feature purification and feature enhancement can purify the cross modal features of color modal and depth modal, and obtain the combined feature $f_{cm}^{i}$ $(i=1,2,3,4,5)$. Specifically, the purification process can be defined as:
\begin{equation}
  \widetilde{f^{i}} = f_{rgb}^{i} \otimes f_{d}^{i},
\end{equation}
\begin{equation}
  \overline{f_{rgb}^{i}} = ((\widetilde{f^{i}} \oplus f_{rgb}^{i}) \otimes f_{rgb}^{i}),
\end{equation}
\begin{equation}
  \overline{f_{d}^{i}} = ((\widetilde{f^{i}} \oplus f_{d}^{i}) \otimes f_{d}^{i}),
\end{equation}
\begin{equation}
  \widehat{f^{i}} = \overline{f_{rgb}^{i}} \oplus \overline{f_{d}^{i}},
\end{equation}
where $\otimes$ and $\oplus$ denote the operations of element-wise multiplication and element-wise addition, respectively. The purification process continues to cross supplement the color features and depth features, purify the depth features with the color features, and then purify the color features with the depth features, so as to iterate repeatedly. In addition, he enhancement process can be defined as:
\begin{equation}
  f_{cm}^{i} = SA(CA(\widehat{f}^{i})),
\end{equation}
where $CA(\cdot)$ and $SA(\cdot)$ denote channel attention mechanism and spatial attention mechanism, respectively. The enhancement process feeds into the channel attention mechanism and spatial attention mechanism in turn. It first enhances the features in the channel way, and then enhances the features in the spatial way.

\subsection{Scale Unification Module}

In the network, the features of each scale have different resolutions due to several sampling. Apart from that, the semantic information and detail information contained in the features of each scale are very different. On the one hand, features of different sizes can not be properly handled in a custom enclosure. On the other hand, feature interaction at all scales is also very important from beginning to end. In order to solve these problems, a Scale Unification Module (SUM) is designed.

\begin{figure}[t] \small
  \centering
  \includegraphics[width=0.37\textwidth]{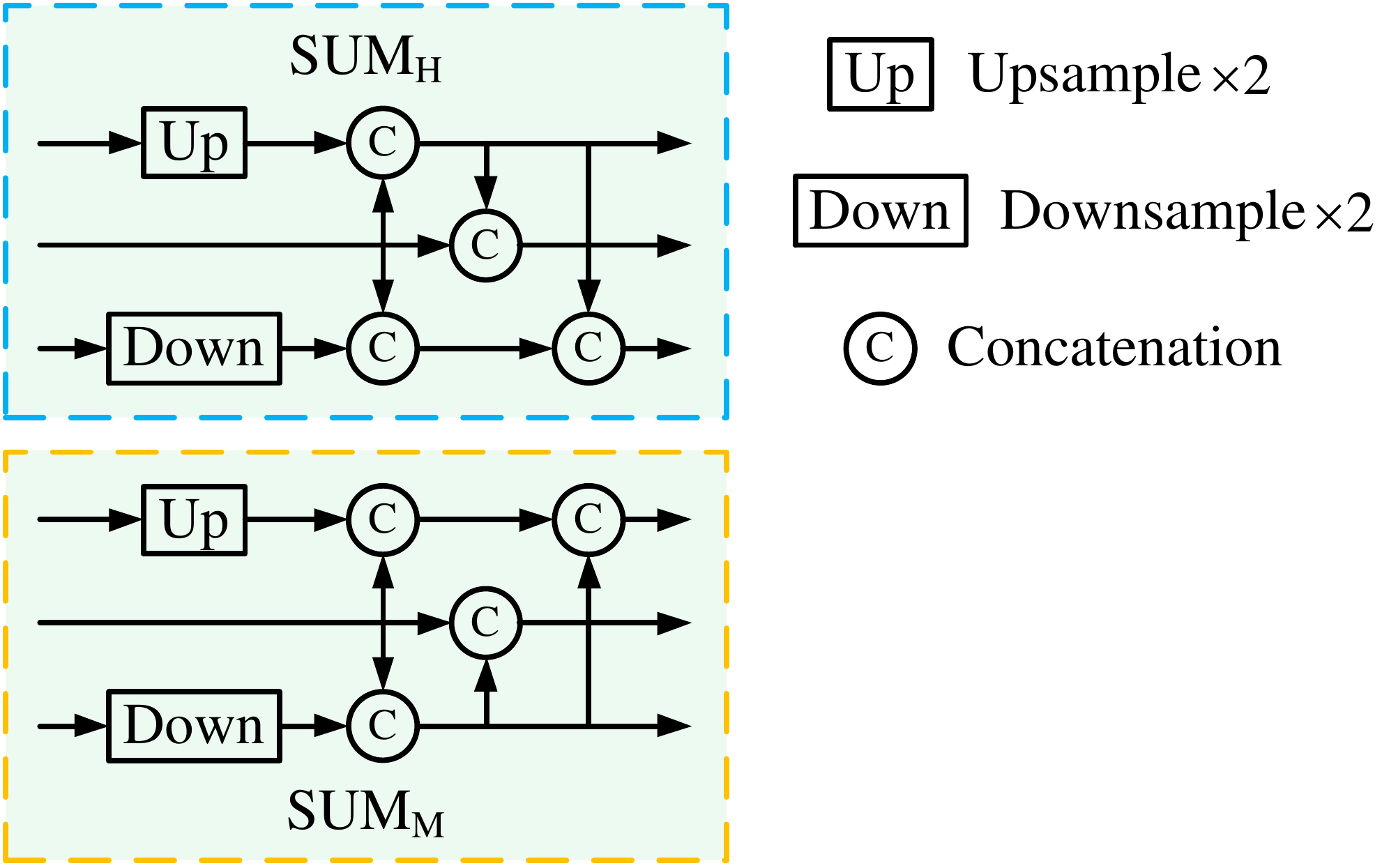} \\
  \caption{The schematic diagram of Scale Unification Module (SUM).}
  \label{Figure 3}
\end{figure}

SUM has two versions corresponding to the two groups formed by soft partition, $SUM_{H}$ and $SUM_{M}$. Their schematic diagram is shown in Figure \ref{Figure 3}. Both of them upsampling or downsampling the higher and lower layers to the scale of the middle layer, respectively, and then a series of symmetrical concatenation operations are carried out.

Before that, the features of all layers $f_{cm}^{i}$ $(i=1,2,3,4,5)$ are passed through a transition layer, so that the number of channels from low to high is collectively normalized from 64, 256, 512, 1024 and 2048 to 64, and the transited feature $f_{t}^{i}$ $(i=1,2,3,4,5)$ is obtained. The transition layer contains a 3$\times$3 convolution layer and a ReLU activation function. The features of the higher three layers and the middle three layers are sent to SUM in block shape, in which the features of the higher three layers $f_{t}^{i}$ $(i=3,4,5)$ are sent to $SUM_{H}$ and the features of the middle three layers $f_{t}^{i}$ $(i=2,3,4)$ are sent to $SUM_{M}$ to obtain the features $hf^{i}$ $(i=3,4,5)$ and $mf^{i}$ $(i=2,3,4)$, respectively.

Let both of the three features of the higher three layers and the middle three layers from high to low be expressed as $f^{h}$, $f^{m}$ and $f^{l}$, respectively. $SUM_{H}$ and $SUM_{M}$ can be defined as
\begin{align}
  &f^{h} = Concat(Up(f^{h}), f^{m}), \\
  &f^{m} = Concat(f^{m}, f^{h}), \\
  &f^{l} = Concat(Concat(Down(f^{l}), f^{m}), f^{h})
\end{align}
and
\begin{align}
  &f^{h} = Concat(Concat(Up(f^{h}), f^{m}), f^{l}), \\
  &f^{m} = Concat(f^{m}, f^{l}), \\
  &f^{l} = Concat(Down(f^{l}), f^{m})
\end{align}
where $Up(\cdot)$ and $Down(\cdot)$ denote the upsampling and downsampling operations, respectively, $Concat(\cdot)$ denotes the concatenation operation.

\subsection{Multiple Transformer Encoder}

In order to make better use of the advantages of transformer, a Multiple Transformer Encoder (MTE) is developed. The encoder includes multiple grouped encoder bodies, and each encoder body is built by several transformer encoders sharing weight. It can learn common information from multi-level features. More importantly, it enjoys high cohesion inside different groups and low coupling outside different groups, which makes the learned feature information more discriminative.

First, the feature $X\in\mathbb{R}^{H\times W}$ is reshaped into a 2D sequence of flattened patches, i.e., $X^{i}\in\mathbb{R}^{P^{2}}$ $(i=1,2,\cdots,N)$, in which $N=(H\times W)/P^{2}$ is the number of patches. In this way, the size of each patch is $P\times P$. Then, a trainable linear projection is used to map the patch to the potential D-dimensional embedding space. After that, the positional information of each patch is encoded through learning position embedding and added to patch embedding. The process can be formalized as
\begin{equation}
  z_{0} = [X^{1}E; X^{2}E; \cdots; X^NE] + E_{pos},
\end{equation}
where $E\in\mathbb{R}^{P^{2}\times D}$ denotes the patch embedding operation, $E_{pos}\in\mathbb{R}^{N\times D}$ denotes the position embedding operation.

The standard transformer encoder is composed of $L$ transformer layers assembled alternately. In each layer, Multihead Self-Attention (MSA) and Multi-Layer Perceptron (MLP) are two key blocks. The layer norm is embedded before them and the residual connection is embedded after them. The specific process is formalized as
\begin{align}
  &z_l^{\prime} = MSA(LN(z_{l-1}))+z_{l-1}, \ l=1,2,\cdots,L, \\
  &z_l = MLP(LN(z_l))+z_l^{\prime}, \ l=1,2,\cdots,L,
\end{align}
where $LN(\cdot)$ denotes the layer norm operation.

In contrast, Multiple Transformer Encoder find another way to achieve the synchronous improvement of encoder and encoder by placing different shared weights between two groups of encoders composed of three standard transformer encoders. What is more, the input sizes of the two groups of encoder bodies are different. The input of $MTE_{H}$ is 16$\times$16. The input of $MTE_{M}$ is 32$\times$32. Their outputs are consistent with their respective inputs. It should be noted that 64$\times$64 is not used as input, because the memory to bear is too large, which is generally difficult to achieve.

\begin{figure}[t] \small
  \centering
  \includegraphics[width=0.42\textwidth]{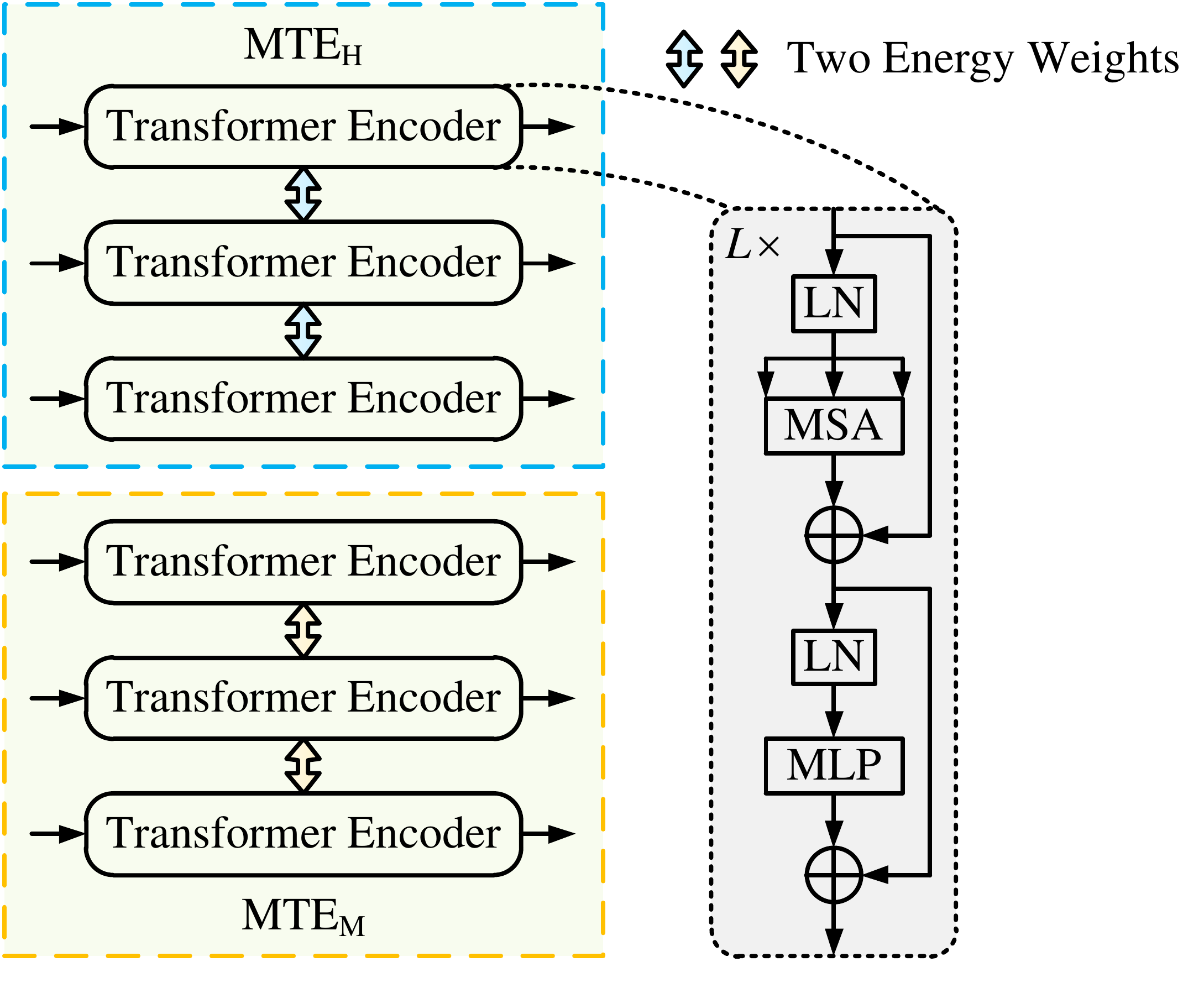} \\
  \caption{The schematic diagram of Multiple Transformer Encoder (MTE).}
  \label{Figure 4}
\end{figure}

MTE has two versions corresponding to the two groups formed by soft partition, $MTE_{H}$ and $MTE_{M}$. Their schematic diagram is shown in Figure \ref{Figure 4}. The transformer encoders used by the two have no other difference except the size of input and output along with the shared energy weight.

The specific processes of $MTE_{H}$ and $MTE_{M}$ can be defined as
\begin{equation}
  h^{\prime} f^{i} = MTE_{H}(hf^{i}),
\end{equation}
and
\begin{equation}
  m^{\prime} f^{i} = MTE_{M}(mf^{i}),
\end{equation}

The features $hf^{i}$ $(i=5,4,3)$ and $mf^{i}$ $(i=4,3,2)$ are passed through two encoders groups with different grid sizes to obtain new features $h^{\prime} f^{i}$ $(i=5,4,3)$ and $m^{\prime} f^{i}$ $(i=4,3,2)$. The two encoders groups share their own weights, and the internal relevance of these features at the high and middle level groups is guaranteed. Meanwhile, the two encoders are irrelevant to each other, which also ensures the external differences of these features at the high and middle level groups.

\subsection{Cluster Integration Module}

In order to cascade the features of different layers and even those with recognition in the same layer in an orderly manner, a Cluster Integration Unit (CIU) is developed.

\begin{figure}[t] \small
  \centering
  \includegraphics[width=0.4\textwidth]{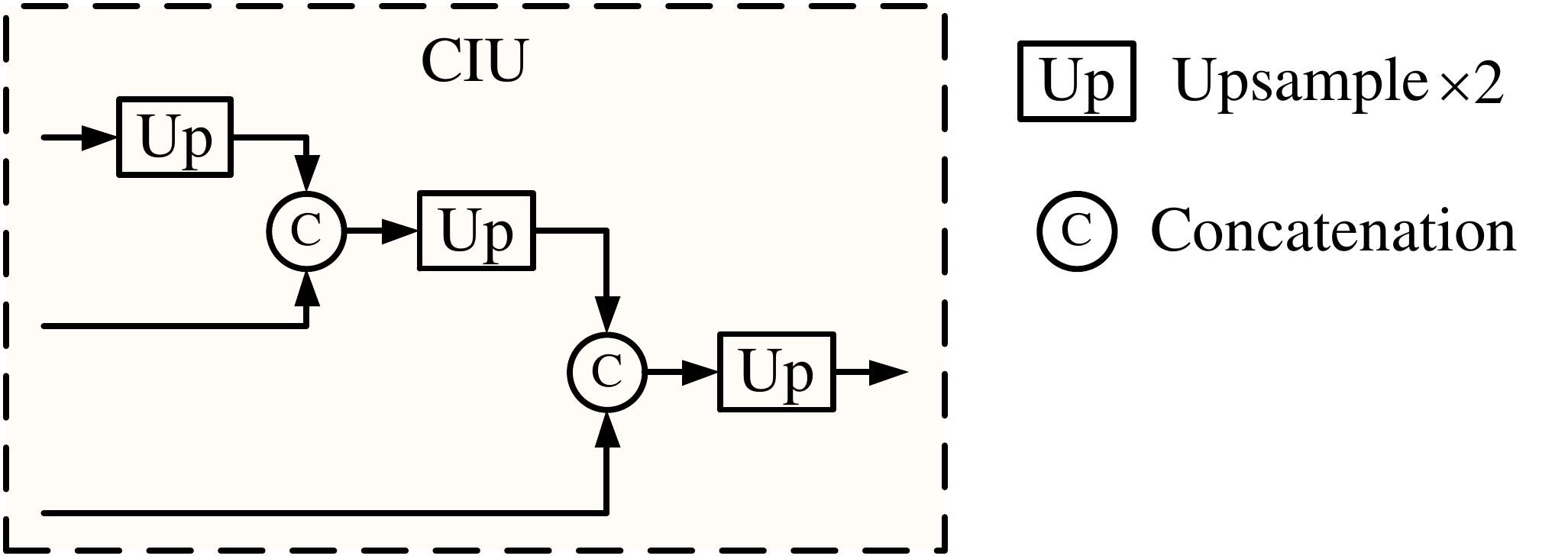} \\
  \caption{The schematic diagram of Cluster Integration Unit (CIU).}
  \label{Figure 5}
\end{figure}

The schematic diagram of CIU is shown in Figure \ref{Figure 5}. It is a cascading way to integrate the clustered features in the order from high to low, which is first upsampled by the high-level features, and then concatenated with the low-level features.

According to the criteria of interleaving difference of features from different encoder bodies, the generated features $\{h^{\prime}f^{5}, h^{\prime}f^{4}, h^{\prime}f^{3}\}$ and $\{m^{\prime}f^{4}, m^{\prime}f^{3}, m^{\prime}f^{2}\}$ are clustered into three classes, namely $C_1 = \{h^{\prime} f^{5}, m^{\prime} f^{4}\}$, $C_2 = \{h^{\prime} f^{4}, m^{\prime} f^{3}\}$ and $C_3 = \{h^{\prime} f^{3}, m^{\prime} f^{2}\}$. In the first class $C_{1}$, $h^{\prime} f^{5}$ contains the feature information of equal fourth and fifth layers, while $m^{\prime} f^{4}$ contains the feature information of equal second and fourth layers and more third layers. Therefore, it has rich information from the second layer to the fifth layer, and this property is suitable for biased lossless features. Similarly, in the second class $C_{2}$, $h^{\prime} f^{4}$ contains the feature information of equal fifth layer and more fourth layer, while $m^{\prime} f^{3}$ contains the feature information of equal second layer and more third layer. At the same time, in the third class $C_{3}$, $h^{\prime} f^{3}$ contains the feature information of equal third layer and fifth layer and more fourth layer, $m^{\prime} f^{2}$ contains equal feature information of the second and third layers. Therefore, they also satisfy the same properties as the first class.

These three classes pay attention to the feature information except the first layer with different emphasis, so it is necessary to combine them with the feature $f_{t}^{1}$ containing the first layer information, respectively. The integration process of all features to be combined can be defined as
\begin{equation}
  f_{i}^{\prime} = Up(Concat(Up(Concat(Up(C_{i1}), C_{i2})), f_{t}^{1})),
\end{equation}
where $C_{ij}$ indicates the $j$-th element in the $i$-th class.

\subsection{Loss Function}
Assuming that there are three side outputs $f_{i}^{\prime}$ $(i=1,2,3)$, the predicted saliency map $S_{i}$ $(i=1,2,3)$ from them is produced as
\begin{equation}
  S_{i} = Sig(Up(f_{i}^{\prime})),
\end{equation}
where $Sig(\cdot)$ denotes the Sigmoid function, which is used to compute the probability for each pixel of being salient or not. It also involves the combination of convolution and ReLU operations.

Similar to \cite{Liu2021TriTransNet}, the supervised learning of error also adopt the Pixel Position Aware (PPA) loss \cite{Wei2020F^3Net}, denoted by $\mathcal{L}_{ppa}$, which is the joint loss function by employing the weighted Binary Cross Entropy (wBCE) loss and weighted Intersection over Union (wIoU) loss.

Thus, the total loss function is defined as
\begin{equation}
  \mathcal{L}_{Total} = \sum_{i=1}^{3} \mathcal{L}_{ppa}(S_{i}, G),
\end{equation}
where $G$ is ground-truth saliency map.

\begin{table*}[!t]
\centering
\small
\renewcommand{\arraystretch}{1.5}
\renewcommand{\tabcolsep}{0.8mm}
\caption{\small Performance comparison with 18 state-of-the-art methods on six datasets. $\uparrow$ and $\downarrow$ stand for larger and smaller is better, respectively. The best results are marked in \textbf{bold}. The \textit{corner note} of each method is the publication year.}
\label{Table 1}
\resizebox{1\textwidth}{!}{
  \begin{tabular}{c|cccc|cccc|cccc|cccc|cccc|cccc}
  \bottomrule[1pt]
  \multirow{2}{*}{\normalsize{Methods}}
  &\multicolumn{4}{c|}{STEREO \cite{Niu2012Leveraging}} &\multicolumn{4}{c|}{NJU2K \cite{Ju2014Depth}} &\multicolumn{4}{c|}{NLPR \cite{Peng2014RGBD}} &\multicolumn{4}{c|}{RGBD135 \cite{Cheng2014Depth}} &\multicolumn{4}{c|}{DUT-RGBD \cite{Piao2019Depth-induced}} &\multicolumn{4}{c}{SIP \cite{Fan2021Rethinking}} \\
  \cline{2-25}
  &$S_{\alpha}\uparrow$ &$F_{\beta}\uparrow$ &$E_{\xi}\uparrow$ &$\mathcal{M}\downarrow$
  &$S_{\alpha}\uparrow$ &$F_{\beta}\uparrow$ &$E_{\xi}\uparrow$ &$\mathcal{M}\downarrow$
  &$S_{\alpha}\uparrow$ &$F_{\beta}\uparrow$ &$E_{\xi}\uparrow$ &$\mathcal{M}\downarrow$
  &$S_{\alpha}\uparrow$ &$F_{\beta}\uparrow$ &$E_{\xi}\uparrow$ &$\mathcal{M}\downarrow$
  &$S_{\alpha}\uparrow$ &$F_{\beta}\uparrow$ &$E_{\xi}\uparrow$ &$\mathcal{M}\downarrow$
  &$S_{\alpha}\uparrow$ &$F_{\beta}\uparrow$ &$E_{\xi}\uparrow$ &$\mathcal{M}\downarrow$ \\
  \hline
  DF$_{17}$ \cite{Qu2017RGBD} &.757 &.742 &.838 &.141 &.735 &.744 &.818 &.151 &.769 &.682 &.838 &.099 &.681 &.573 &.806 &.132 &.719 &.748 &.842 &.150 &.653 &.673 &.794 &.185 \\
  CTMF$_{18}$ \cite{Han2018CNNs-based} &.848 &.771 &.870 &.086 &.849 &.779 &.864 &.085 &.860 &.723 &.869 &.056 &.863 &.778 &.911 &.055 &.830 &.790 &.882 &.097 &.716 &.684 &.824 &.139 \\
  MMCI$_{19}$ \cite{Chen2019Multi-modal} &.873 &.829 &.905 &.068 &.859 &.803 &.878 &.079 &.856 &.729 &.871 &.856 &.848 &.762 &.904 &.065 &.791 &.751 &.855 &.113 &.833 &.795 &.886 &.086 \\
  TANet$_{19}$ \cite{Chen2019Three-stream} &.871 &.835 &.916 &.060 &.878 &.844 &.909 &.061 &.886 &.795 &.916 &.041 &.858 &.795 &.919 &.046 &.808 &.771 &.866 &.093 &.835 &.809 &.894 &.075 \\
  DMRA$_{19}$ \cite{Piao2019Depth-induced} &.752 &.762 &.816 &.086 &.886 &.872 &.921 &.051 &.899 &.855 &.942 &.031 &.900 &.866 &.944 &.030 &.888 &.883 &.930 &.048 &.800 &.815 &.858 &.088 \\
  ICNet$_{20}$ \cite{Li2020ICNet} &.903 &.865 &.915 &.045 &.894 &.868 &.905 &.052 &.923 &.870 &.944 &.028 &.920 &.889 &.959 &.027 &.852 &.830 &.897 &.072 &.854 &.836 &.899 &.069 \\
  DCMF$_{20}$ \cite{Chen2020RGBD} &.883 &.841 &.904 &.054 &.889 &.859 &.897 &.052 &.900 &.839 &.933 &.035 &.877 &.820 &.923 &.040 &.798 &.750 &.848 &.104 &.859 &.819 &.898 &.068 \\
  CoNet$_{20}$ \cite{Ji2020Accurate} &.905 &.884 &.927 &.037 &.895 &.872 &.912 &.046 &.908 &.846 &.934 &.031 &.911 &.861 &.945 &.027 &.919 &.909 &.948 &.033 &.858 &.842 &.909 &.063 \\
  DANet$_{20}$ \cite{Zhao2020A} &.901 &.868 &.921 &.043 &.899 &.871 &.908 &.045 &.920 &.875 &.951 &.027 &.924 &.899 &.968 &.023 &.899 &.888 &.934 &.043 &.875 &.855 &.914 &.054 \\
  JL-DCF$_{20}$ \cite{Fu2020JL-DCF} &.903 &.869 &.919 &.040 &.902 &.885 &.913 &.041 &.925 &.878 &.953 &.022 &.931 &.900 &.969 &.020 &.906 &.882 &.931 &.043 &.880 &.873 &.921 &.049 \\
  SSF$_{20}$ \cite{Zhang2020Select} &.887 &.867 &.921 &.046 &.899 &.886 &.913 &.043 &.914 &.875 &.949 &.026 &.905 &.876 &.948 &.025 &.916 &.914 &.946 &.034 &.868 &.851 &.911 &.056 \\
  UC-Net$_{20}$ \cite{Zhang2020UC-Net} &.903 &.885 &.922 &.039 &.897 &.889 &.903 &.043 &.920 &.890 &.953 &.025 &.933 &.917 &.974 &.018 &.864 &.856 &.903 &.056 &.875 &.868 &.913 &.051 \\
  A2dele$_{20}$ \cite{Piao2020A2dele} &.878 &.874 &.915 &.044 &.869 &.874 &.897 &.051 &.896 &.878 &.945 &.028 &.885 &.865 &.922 &.028 &.886 &.890 &.924 &.043 &.826 &.825 &.892 &.070 \\
  S$^2$MA$_{20}$ \cite{Liu2020Learning} &.890 &.855 &.907 &.051 &.894 &.865 &.896 &.053 &.915 &.853 &.938 &.030 &.941 &.906 &.974 &.021 &.903 &.866 &.921 &.044 &.872 &.854 &.911 &.057 \\
  D$^3$Net$_{21}$ \cite{Fan2021Rethinking} &.899 &.859 &.920 &.046 &.901 &.865 &.914 &.046 &.912 &.861 &.944 &.030 &.898 &.870 &.951 &.031 &.775 &.756 &.847 &.097 &.860 &.835 &.902 &.063 \\
  CDNet$_{21}$ \cite{Jin2021CDNet} &.896 &.873 &.922 &.042 &.885 &.866 &.911 &.048 &.902 &.848 &.935 &.032 &.875 &.839 &.921 &.034 &.880 &.874 &.918 &.048 &.823 &.805 &.880 &.076 \\
  HAINet$_{21}$ \cite{Li2021Hierarchical} &.907 &.885 &.925 &.040 &.912 &.900 &.922 &.038 &.924 &.897 &.957 &.024 &.935 &.924 &.974 &.018 &.910 &.906 &.938 &.038 &.880 &.875 &.919 &.053 \\
  TriTransNet$_{21}$ \cite{Liu2021TriTransNet} &\textbf{.908} &.893 &.927 &.033 &.920 &.919 &.925 &.030 &\textbf{.928} &\textbf{.909} &.960 &.020 &.943 &.936 &.981 &\textbf{.014} &.933 &.938 &.957 &.025 &.886 &.892 &.924 &.043 \\
  \hline
  GroupTransNet (ours) &\textbf{.908} &\textbf{.895} &\textbf{.928} &\textbf{.032}
       &\textbf{.922} &\textbf{.921} &\textbf{.926} &\textbf{.028}
       &\textbf{.928} &.908 &\textbf{.961} &\textbf{.019}
       &\textbf{.944} &\textbf{.938} &\textbf{.982} &\textbf{.014}
       &\textbf{.935} &\textbf{.939} &\textbf{.958} &\textbf{.024}
       &\textbf{.887} &\textbf{.895} &\textbf{.926} &\textbf{.041} \\
  \toprule[1pt]
  \end{tabular}}
\end{table*}

\section{Experiments}

In this section, we first describe the commonly used datasets and evaluation metrics, and then elaborate the specific training protocols and implementation details. Next, we quantitatively and qualitatively compare the proposed method with the state-of-the-art RGB-D salient object detection methods to prove its advantages. Afterwards, we conduct a series of ablation studies to verify the role of each component of our network. In addition, we also discuss some failure cases of the model.


\begin{figure*}[ht] \small
  \centering
  \includegraphics[width=1\textwidth]{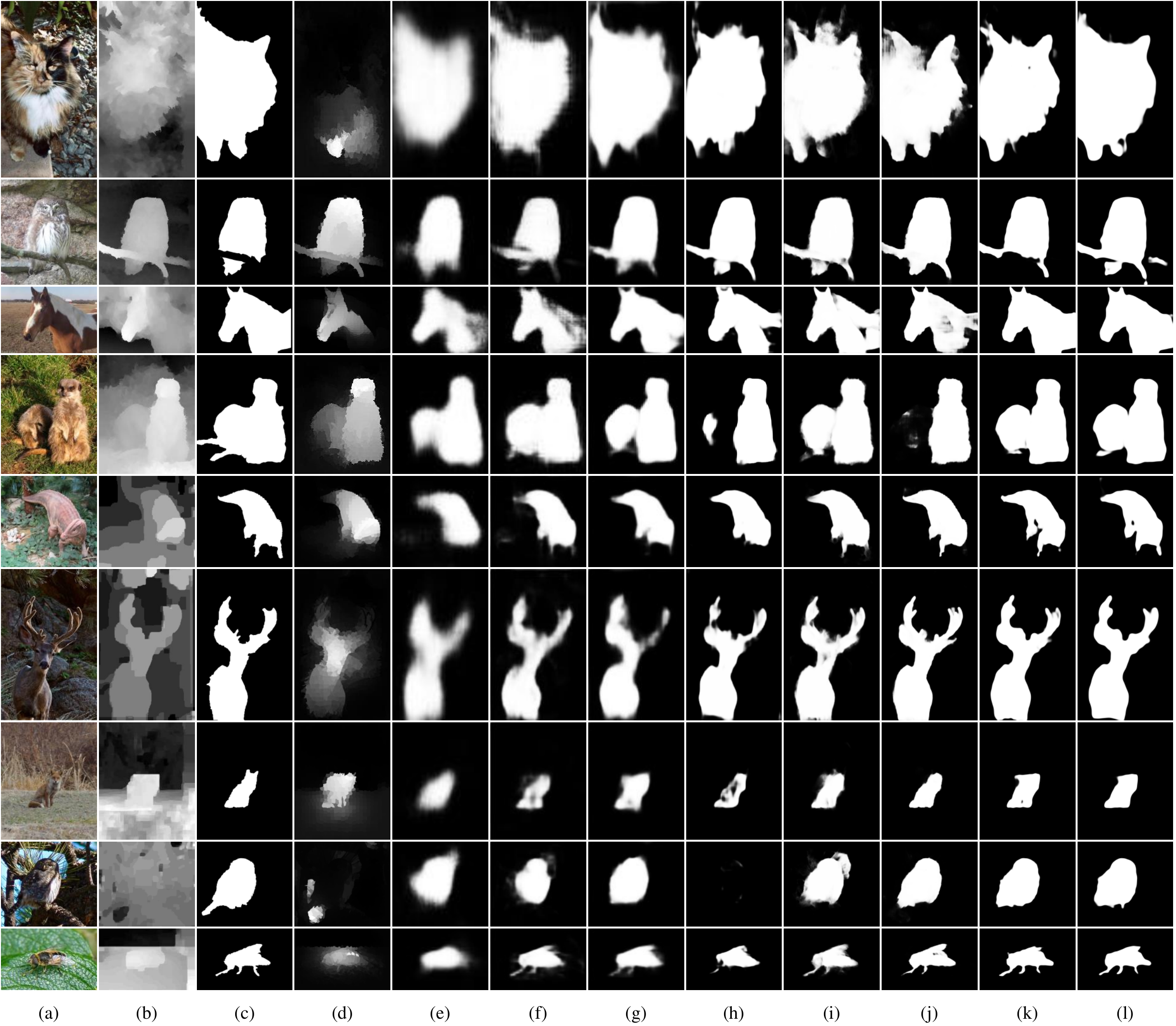} \\
  \caption{Comparison of visualization results. (a) RGB; (b) Depth; (c) Ground-truth; (d) DF; (e) CTMF; (f) MMCI; (g) TANet; (h) DMRA; (i) D$^3$Net; (j) HAINet; (k) TriTransNet; (l) The proposed method.}
  \label{Figure 7}
\end{figure*}

\subsection{Experimental Setup}

\subsubsection{Datasets}
To measure the performance of the proposed method, we execute validation on six benchmark datasets, including STEREO \cite{Niu2012Leveraging}, NJU2K \cite{Ju2014Depth}, NLPR \cite{Peng2014RGBD}, RGBD135 \cite{Cheng2014Depth}, DUT-RGBD \cite{Piao2019Depth-induced} and SIP \cite{Fan2021Rethinking}.

STEREO, also called SSB1000, is composed of 1000 stereoscopic image pairs and corresponding ground-truth downloaded from the Internet, in which the depth images are estimated from the stereo images.

NJU2K consists of 1985 image pairs along with ground-truth collected from the Internet, 3D movies and photographs, in which the depth images are estimated from the stereo images.

NLPR contains 1000 image pairs with their ground-truth taken under different illumination conditions, in which the depth images are captured by the Microsoft Kinect.

RGBD135, also known as DES, is composed of 135 indoor image pairs and the ground-truth, in which the depth maps are captured by the Microsoft Kinect.


DUT-RGBD contains 1200 image pairs with their ground-truth taken in varied real life situations, in which the depth images are captured by the Lytro light field camera.

SIP has 929 high-resolution image pairs and corresponding ground-truth of person activities, in which the depth images are captured by the Huawei Meta10.

\subsubsection{Evaluation Metrics}
To measure the performance of the proposed method and other methods, we choose four typical evaluation metrics, including S-measure ($S_{\alpha}$) \cite{Fan2017Structure-measure}, F-measure ($F_\beta$) \cite{Achanta2009Frequency-tuned}, Mean Absolute Error that is abbreviated as MAE ($\mathcal{M}$) \cite{Perazzi2012Saliency} and E-measure ($E_{\xi}$) \cite{Fan2018Enhanced-alignment}.

S-measure calculates the structural similarity between the prediction and ground-truth, which is defined as
\begin{equation}
  S_{\alpha} = \alpha\times S_o + (1-\alpha) \times S_r,
\end{equation}
where $S_o$ and $S_r$ denote the object-aware structural similarity and region-aware structural similarity, respectively, and $\alpha$ is set to 0.5 to assign equal constraints to both.

F-measure calculates the harmonic mean of the thresholded precision and recall, which is defined as
\begin{equation}
  F_\beta = \frac{(1+\beta^2)\times Precision\times Recall}{\beta^2\times Precision + Recall},
\end{equation}
where $\beta^2$ is set to 0.3 to emphasize precision more than recall as suggested in \cite{Borji2015Salient}. We adopts the average F-measure.

E-measure calculates the local pixel-level error and the global image-level error together, which is defined as
\begin{equation}
  E_{\xi} = \frac{1}{W\times H}\sum _{i=1}^{W}\sum _{j=1}^{H}\phi(x,y),
\end{equation}
where $\phi$ denote the enhanced alignment matrix, $W$ and $H$ are the width and height of the image, respectively.

MAE calculates the average pixel-wise absolute difference between the prediction and ground-truth, which is defined as
\begin{equation}
  \mathcal{M} = \frac{1}{W\times H}\sum _{i=1}^{W}\sum _{j=1}^{H}|P(i,j)-G(i,j)|,
\end{equation}
where
$P$ denote the prediction, $G$ denote the ground-truth.

Moreover, we also plot the Precision-Recall (PR) curve and F-measure ($F_\beta$) curve.

\subsubsection{Training Protocols}
For the sake of fairness, we follow the same mainstream protocols just like the previous works \cite{Chen2020Progressively, Fan2021Rethinking} to employ the first training setting relative to STEREO, NJU2K, NLPR, RGBD135 and SIP, and as in the previous works \cite{Piao2019Depth-induced, Ji2020Accurate, Piao2020A2dele} to employ the second training setting relative to DUT-RGBD. When the validation is executed on the former, 1485 image pairs from NJU2K and 700 image pairs from NLPR are split for training, the remaining samples of these two and the entire samples of the other three are collected for testing. When the validation is executed on the latter, additional 800 image pairs from DUT-RGBD are split for training, and the remaining samples of it are collected for testing.

\subsubsection{Implementation Details}
We implement the experiment on four NVIDIA GeForce RTX 3090 GPUs with 24GB memory using the PyTorch toolbox. The ResNet-50 \cite{He2016Deep} pretrained on ImageNet \cite{Deng2009ImageNet} is used as the backbone.
To avoid potential overfitting, image pairs in training set are augmented with random flipping, cropping and rotating. During both training and testing stages, the resolution of all RGB and depth input image pairs are resized to 256$\times$256.
Our network is end-to-end and does not use any pre-processing (e.g., HHA \cite{Gupta2014Learning}) or post-processing (e.g., CRF \cite{Krahenbuhl2011Efficient}).
The Adam \cite{Kingma2014Adam} optimizer is adopted to optimize the network. The initial learning rate is set to 1e-4, which is adjusted by multiplying by 0.1 after every 60 epochs.
Our model is trained for 150 epochs with a batch size of 3.
It takes about 15 hours to train the model.
The test time for each image pair is only 0.037 seconds.

\subsection{Comparison With the State-of-the-Arts}
We compare the proposed method with 18 state-of-the-art methods, including DF \cite{Qu2017RGBD}, CTMF \cite{Han2018CNNs-based}, MMCI \cite{Chen2019Multi-modal}, TANet \cite{Chen2019Three-stream}, DMRA \cite{Piao2019Depth-induced}, ICNet \cite{Li2020ICNet}, DCMF \cite{Chen2020RGBD}, CoNet \cite{Ji2020Accurate}, DANet \cite{Zhao2020A}, JL-DCF \cite{Fu2020JL-DCF}, SSF \cite{Zhang2020Select}, UC-Net \cite{Zhang2020UC-Net}, A2dele \cite{Piao2020A2dele}, S$^2$MA \cite{Liu2020Learning}, D$^3$Net \cite{Fan2021Rethinking}, CDNet \cite{Jin2021CDNet}, HAINet \cite{Li2021Hierarchical} and TriTransNet \cite{Liu2021TriTransNet}. The saliency maps of these competitive methods are provided by the corresponding authors or generated by running their released source codes under the default parameters.

\subsubsection{Quantitative Comparison}
The results of quantitative comparison are listed in Table. \ref{Table 1}.
We can see that our GroupTransNet is superior to competitors in almost all metrics.
Compared with the second best method TriTransNet, which also uses transformer, the proposed method averagely achieves 0.6\%, 0.9\% and 0.7\% performance gains on all datasets in terms of metrics of $S_{\alpha}$, $F_{\beta}$ and $E_{\xi}$, respectively. In particular, the performance of this method compared with the second best method TriTransNet on NJU2K and SIP datasets in terms of metric of $\mathcal{M}$ is improved by 0.2\%, respectively.
In addition, the method with transformer has significantly better performance than the method without transformer.

\subsubsection{Qualitative Comparison}
The results of qualitative comparison are shown in Figure. \ref{Figure 7}. As can be seen from it, GroupTransNet has the ability to accurately identify salient objects in a variety of scenarios, such as similar foreground and background, small objects, multiple objects, etc.
In particular, TriTransNet, HAINet and CDNet may not highlight the object object well, but the proposed method can highlight the clear structure of the whole completely and accurately, and maintain a high degree of similarity with the real results.

\subsection{Ablation Analysis}
To demonstrate the impact of each component of our proposed network on the performance of salient object detection, we carry out experiments on NJU2K and SIP, involving four ablation studies.
\subsubsection{Effect of MPM}
Table. \ref{Table 2} lists the results of ablation studies for MPM.
The scheme of `No.1' used here is baseline, which means that MPM is not applied.
There are two other modules that are somewhat similar to MPM, namely, DEM \cite{Fan2020BBS-Net} and DPM \cite{Liu2021TriTransNet}.
The schemes of `No.2', `No.3' and `No.4' represent adding them to the baseline, respectively.
It can be seen that the embedding of MPM has a very significant effect on the improvement of performance. In addition, the performance improvement brought by the embedding of MPM is better than that of DEM or DPM.
\begin{table}[t!]
\centering
\small
\renewcommand{\arraystretch}{1.5}
\renewcommand{\tabcolsep}{0.8mm}
\caption{\small Ablation studies of MPM on NJU2K and SIP. The best result of each column is \textbf{bold}.}
\label{Table 2}
\resizebox{0.49\textwidth}{!}{
  \begin{tabular}{c|cccc|cccc|cccc}
  \bottomrule[1pt]
  \multirow{2}{*}{No.}
  &\multirow{2}{*}{Baseline}
  &\multirow{2}{*}{DEM}
  &\multirow{2}{*}{DPM}
  &\multirow{2}{*}{MPM}
  &\multicolumn{4}{c|}{NJU2K} &\multicolumn{4}{c}{SIP} \\
  \cline{6-13}
  & & & &
  &$S_{\alpha}\uparrow$ &$F_{\beta}\uparrow$ &$E_{\xi}\uparrow$ &$\mathcal{M}\downarrow$ &$S_{\alpha}\uparrow$ &$F_{\beta}\uparrow$ &$E_{\xi}\uparrow$ &$\mathcal{M}\downarrow$ \\
  \hline
  1 &$\surd$ & & & &.909 &.904 &.920 &.035 &.883 &.887 &.921 &.044 \\
  2 &$\surd$ &$\surd$ & & &.914 &.910 &.922 &.033 &.884 &.889 &.923 &.044 \\
  3 &$\surd$ & &$\surd$ & &.920 &.919 &.925 &.030 &.886 &.892 &.924 &.043 \\
  4 &$\surd$ & & &$\surd$ &\textbf{.922} &\textbf{.921} &\textbf{.926} &\textbf{.028} &\textbf{.887} &\textbf{.895} &\textbf{.926} &\textbf{.041} \\
  \toprule[1pt]
  \end{tabular}}
\end{table}
\subsubsection{Effect of SUM}
Table. \ref{Table 3} lists the results of ablation studies for SUM.
Here, the scheme of `Baseline' completely isolates the participation of SUM. Without the participation of SUM$_{H}$ and SUM$_{M}$, they corresponding to the schemes of `w/o SUM$_{H}$' and `w/o SUM$_{M}$'. The scheme of `+SUM' represents the use of SUM.
It can be seen that when removing SUM, it can obtain $\mathcal{M}$ of 0.036 on NJU2K and $\mathcal{M}$ of 0.045 on SIP, respectively. This is continuously improved when SUM$_{H}$ and SUM$_{M}$ are deleted and SUM is used, respectively.
\begin{table}[t!]
\centering
\small
\renewcommand{\arraystretch}{1.5}
\renewcommand{\tabcolsep}{0.8mm}
\caption{\small Ablation studies of SUM on NJU2K and SIP. The best result of each column is \textbf{bold}.}
\label{Table 3}
\resizebox{0.38\textwidth}{!}{
  \begin{tabular}{c|cccc|cccc}
  \bottomrule[1pt]
  \multirow{2}{*}{Variants}
  &\multicolumn{4}{c|}{NJU2K} &\multicolumn{4}{c}{SIP} \\
  \cline{2-9}
  &$S_{\alpha}\uparrow$ &$F_{\beta}\uparrow$ &$E_{\xi}\uparrow$ &$\mathcal{M}\downarrow$
  &$S_{\alpha}\uparrow$ &$F_{\beta}\uparrow$ &$E_{\xi}\uparrow$ &$\mathcal{M}\downarrow$ \\
  \hline
  Baseline &.909 &.906 &.919 &.036 &.881 &.885 &.920 &.045 \\
  \hline
  w/o SUM$_{H}$ &.914 &.910 &.921 &.033 &.882 &.889 &.921 &.044 \\
  w/o SUM$_{M}$ &.917 &.915 &.923 &.030 &.884 &.892 &.923 &.043 \\
  \hline
  +SUM &\textbf{.922} &\textbf{.921} &\textbf{.926} &\textbf{.028} &\textbf{.887} &\textbf{.895} &\textbf{.926} &\textbf{.041} \\
  \toprule[1pt]
  \end{tabular}}
\end{table}
\subsubsection{Effect of MTE}
Table. \ref{Table 4} lists the results of ablation studies for MTE.
Here, the scheme of `Baseline' completely isolates the participation of MTE. Without the participation of MTE$_{H}$ and MTE$_{M}$, they corresponding to the schemes of `w/o MTE$_{H}$' and `w/o MTE$_{M}$'. The scheme of `+MTE' represents the use of MTE.
It can be seen that compared with removing MTE, removing MTE$_{H}$ and MTE$_{M}$ bring 0.4\% and 0.5\% improvement in terms of $\mathcal{M}$ on NJU2K, and both bring 0.4\% improvement in terms of $\mathcal{M}$ on SIP, respectively. When using MTE, the best results can be obtained.
\begin{table}[t!]
\centering
\small
\renewcommand{\arraystretch}{1.5}
\renewcommand{\tabcolsep}{0.8mm}
\caption{\small Ablation studies of MTE on NJU2K and SIP. The best result of each column is \textbf{bold}.}
\label{Table 4}
\resizebox{0.38\textwidth}{!}{
  \begin{tabular}{c|cccc|cccc}
  \bottomrule[1pt]
  \multirow{2}{*}{Variants}
  &\multicolumn{4}{c|}{NJU2K} &\multicolumn{4}{c}{SIP} \\
  \cline{2-9}
  &$S_{\alpha}\uparrow$ &$F_{\beta}\uparrow$ &$E_{\xi}\uparrow$ &$\mathcal{M}\downarrow$
  &$S_{\alpha}\uparrow$ &$F_{\beta}\uparrow$ &$E_{\xi}\uparrow$ &$\mathcal{M}\downarrow$ \\
  \hline
  Baseline &.889 &.887 &.914 &.038 &.868 &.861 &.896 &.050 \\
  \hline
  w/o MTE$_{H}$ &.899 &.898 &.918 &.034 &.871 &.869 &.904 &.046 \\
  w/o MTE$_{M}$ &.904 &.903 &.921 &.033 &.873 &.873 &.908 &.046 \\
  \hline
  +MTE &\textbf{.922} &\textbf{.921} &\textbf{.926} &\textbf{.028} &\textbf{.887} &\textbf{.895} &\textbf{.926} &\textbf{.041} \\
  \toprule[1pt]
  \end{tabular}}
\end{table}
\subsubsection{Effect of CIU}
Table. \ref{Table 5} lists the results of ablation studies for CIU.
The scheme of `No.1' used here is baseline, which means that CIU is not applied.
There are two other options different from CIU in feature fusion details.
When all input features are fused together to produce an output, we name this method as Overall Merger Unit (OMU).
When these input features are fused according to the criteria of non-interleaving difference of features to generate three outputs, we name this method as Corresponding Coupling Unit (CCU).
The schemes of `No.2', `No.3' and `No.4' represent adding them to the baseline, respectively.
It can be seen that with the embedding of OMU, CCU or CIU, the effect on performance improvement is getting better and better.
\begin{table}[t!]
\centering
\small
\renewcommand{\arraystretch}{1.5}
\renewcommand{\tabcolsep}{0.8mm}
\caption{\small Ablation studies of CIU on NJU2K and SIP. The best result of each column is \textbf{bold}.}
\label{Table 5}
\resizebox{0.49\textwidth}{!}{
  \begin{tabular}{c|cccc|cccc|cccc}
  \bottomrule[1pt]
  \multirow{2}{*}{No.}
  &\multirow{2}{*}{Baseline}
  &\multirow{2}{*}{OMU}
  &\multirow{2}{*}{CCU}
  &\multirow{2}{*}{CIU}
  &\multicolumn{4}{c|}{NJU2K} &\multicolumn{4}{c}{SIP} \\
  \cline{6-13}
  & & & &
  &$S_{\alpha}\uparrow$ &$F_{\beta}\uparrow$ &$E_{\xi}\uparrow$ &$\mathcal{M}\downarrow$
  &$S_{\alpha}\uparrow$ &$F_{\beta}\uparrow$ &$E_{\xi}\uparrow$ &$\mathcal{M}\downarrow$ \\
  \hline
  1 &$\surd$ & & & &.915 &.911 &.917 &.032 &.881 &.888 &.917 &.044 \\
  2 &$\surd$ &$\surd$ & & &.917 &.914 &.922 &.032 &.882 &.890 &.919 &.043 \\
  3 &$\surd$ & &$\surd$ & &.919 &.918 &.924 &.030 &.885 &.893 &.924 &.042 \\
  4 &$\surd$ & & &$\surd$ &\textbf{.922} &\textbf{.921} &\textbf{.926} &\textbf{.028} &\textbf{.887} &\textbf{.895} &\textbf{.926} &\textbf{.041} \\
  \toprule[1pt]
  \end{tabular}}
\end{table}

\subsection{Failure Cases}
\begin{figure}[t] \small
  \centering
  \includegraphics[width=0.3\textwidth]{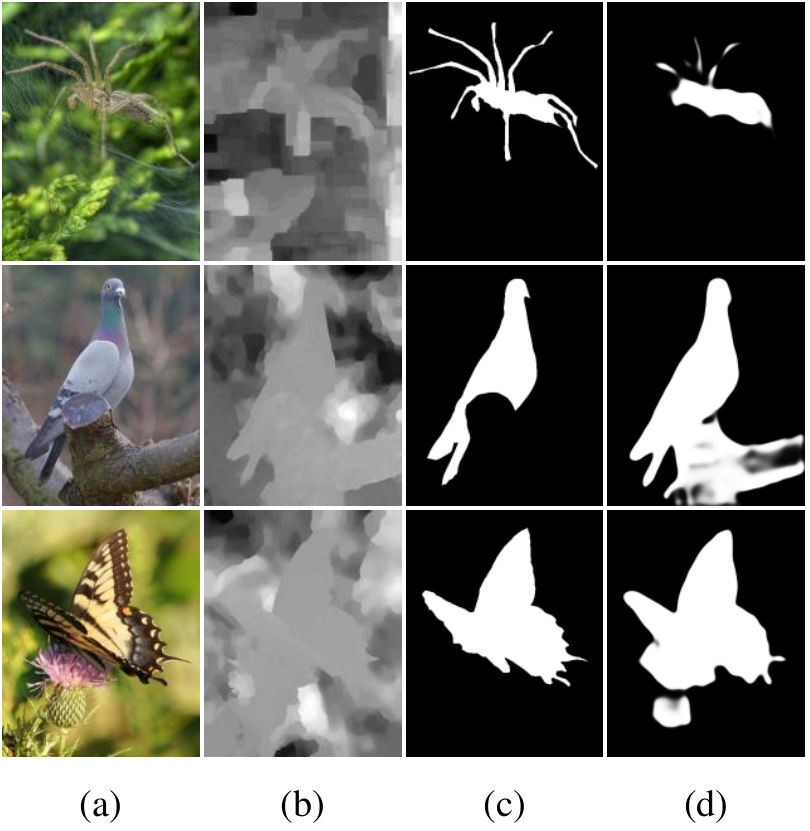} \\
  \caption{Some failure cases. (a) RGB; (b) Depth; (c) Ground-truth; (d) Ours.}
  \label{Figure 8}
\end{figure}

In Figure. \ref{Figure 8}, the failure cases of the proposed method in dealing with detection in some extreme environments are shown.
It can be seen that when the object is in a confusing state in the color map and depth map, or the information of the object in the depth map is not accurate enough, the detection performance of the method will deteriorate more or less.
For example, in first line in the figure, the touch limbs of the spider are blurred in the color map and depth map, which can not be well distinguished. The quality of the image of the second and third lines in the figure is very poor, so that its target is quite different from the shape of the pigeon and butterfly in the real picture.
Therefore, there are some solutions that can be tried. For example, sorting the detected significant objects or intentionally deleting some invalid depth maps.

\section{Conclusion}
In this paper, we propose a novel group transformer network (GroupTransNet) for RGB-D salient object detection. GroupTransNet makes use of transformer to learn the long-range dependencies of cross layer features at the lowest cost, so as to obtain perfect feature expression. This method contains four basic components, namely, Modal Purification Module (MPM), Scale Unification Module (SUM), Multiple Transformer Encoder (MTE) and Cluster Integration Unit (CIU).
Specifically, MPM is used to cross supplement the color modal and depth modal in the purification process, and feed by the channel and spatial way in the enhancement process, so as to realize the purification of cross modal features. Through SUM, the upper and lower layers are respectively upsampled or downsampled to the scale of the middle layer and processed, so as to realize the unity and correlation of the size of each scale feature. Through MTE, different shared energy weights are placed to achieve the effects of high cohesion and low coupling in different groups. And CIU is used to cluster and integrate the features of different layersin a cascade manner.
Compared with 18 state-of-the-art methods, the experimental results on six datasets in terms of four evaluation metrics confirm that the proposed method is superior.

\section*{Acknowledgments}

This work was supported by the National Key R\&D Program of China under Grant 2019YFB1311804, the National Natural Science Foundation of China under Grant 61973173, 91848108 and 91848203, and the Technology Research and Development Program of Tianjin under Grant 18ZXZNGX00340 and 20YFZCSY00830.

\bibliographystyle{IEEEtran}
\bibliography{BibTeX}

\end{document}